\documentclass[11pt,a4paper]{article}
\usepackage[hyperref]{acl2021}
\usepackage{times}
\usepackage{latexsym}
\usepackage{graphicx}
\usepackage{amssymb}
\usepackage{multirow,booktabs}

\usepackage{microtype}

\aclfinalcopy 

\usepackage{amsmath}
\DeclareMathOperator*{\argmax}{argmax}

\usepackage{xcolor}
\definecolor{emerald}{rgb}{0.31, 0.78, 0.47}

\title{GrammarTagger: A Multilingual, Minimally-Supervised \\
Grammar Profiler for Language Education}

\author{Masato Hagiwara \\
  Octanove Labs \\
  Seattle, WA, USA \\
  \texttt{masato@octanove.com} \\\And
  Joshua Tanner\thanks{\ \ Work performed during internship at Octanove Labs.} \\
  University of Washington \\
  University of Tokyo \\
  Tokyo, Japan \\
  \texttt{jotanner@uw.edu} \\\And
  Keisuke Sakaguchi \\
  Allen Institute for AI\\
  Seattle, WA, USA \\
  \texttt{keisukes@allenai.org} \\
  }

\date{}

\begin{document}
\maketitle
\begin{abstract}
We present GrammarTagger, an open-source grammar profiler which, given an input text, identifies grammatical features useful for language education. The model architecture enables it to learn from a small amount of texts annotated with spans and their labels, which 1) enables easier and more intuitive annotation, 2) supports overlapping spans, and 3) is less prone to error propagation, compared to complex hand-crafted rules defined on constituency/dependency parses. We show that we can bootstrap a grammar profiler model with $F_1 \approx 0.6$ from only a couple hundred sentences both in English and Chinese, which can be further boosted via learning a multilingual model. With GrammarTagger, we also build Octanove Learn, a search engine of language learning materials indexed by their reading difficulty and grammatical features\footnote{The code and pretrained models are publicly available at \url{https://github.com/octanove/grammartagger}. See \url{https://www.youtube.com/watch?v=8ujYzoezMhI} for video demonstration.}.
\end{abstract}

\section{Introduction}

Grammar plays an important role in second language (L2) acquisition and education~\cite{long1991focus}, and accurately identifying grammatical features in natural language texts has a wide range of applications, such as highlighting grammatical forms~\cite{meurers-etal-2010-enhancing} and finding authentic materials that match personal interests and proficiency levels~\cite{heilman2008retrieval} for L2 learners and instructors. Grammatical and syntactic features also play an important role in other educational applications, notably readability assessment for L1~\cite{vajjala2012improving} and  L2~\cite{heilman2007combining,xia2016cambridge} acquisition. 

\begin{figure}[!t]
\begin{center}
\includegraphics[scale=0.4]{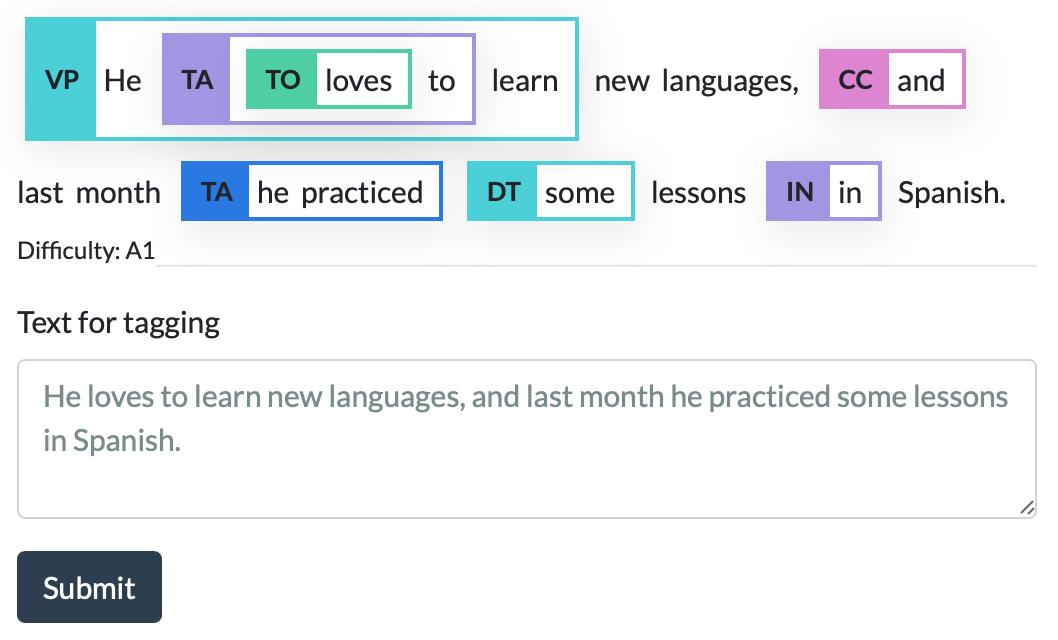} 
\caption{Screenshot of GrammarTagger. GrammarTagger identifies grammatical items in a given text.}
\label{fig:screenshot}
\end{center}
\end{figure}

There have been previous efforts to build grammar-related resources such as English Grammar Profile~\cite{harrison2015english}, A Core Inventory for General English~\cite{north2010eaquals}, and Global Scale of English (GSE) Teacher Toolkit\footnote{\url{https://www.pearson.com/english/about/gse/teacher-toolkit.html}}. However, they are static and require expertise to apply in real-word instructional settings. For example, it is not possible for instructors and learners to choose appropriate learning materials solely based on these resources. In this paper, we address {\it grammatical profiling}, the task of automatically identifying grammatical features called {\it grammatical items} (GIs) contained in a given natural language text.

\begin{figure*}[!t]
\begin{center}
\includegraphics[scale=0.4]{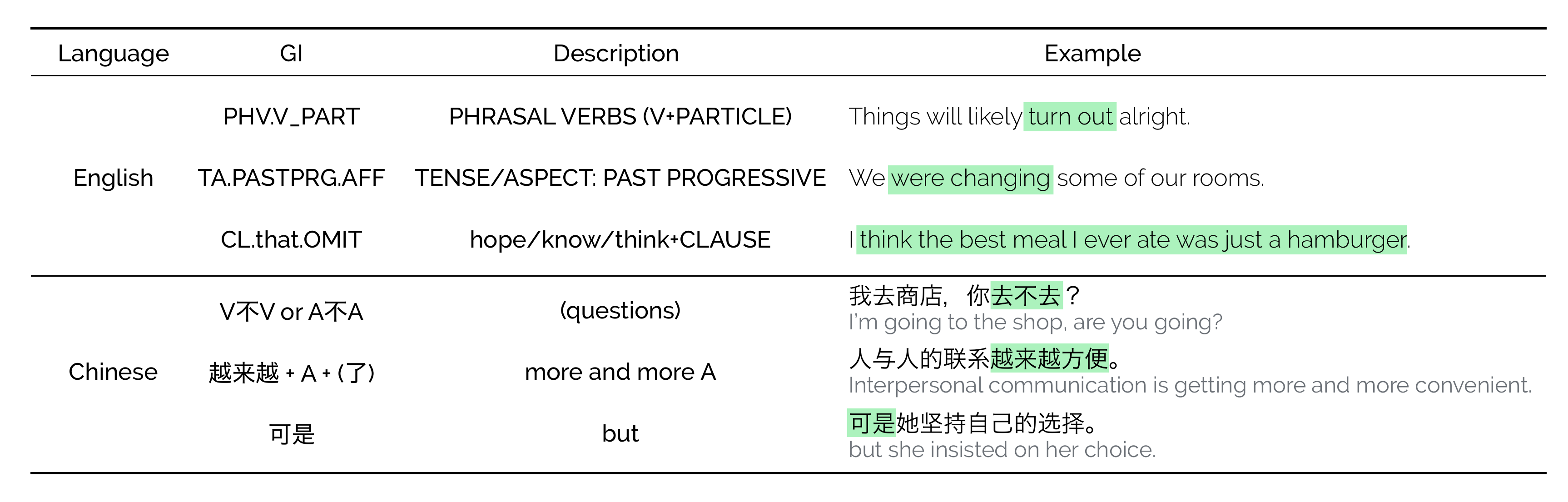} 
\caption{Examples of grammatical items (GIs) in English and Chinese.}
\label{fig:examples}
\end{center}
\end{figure*}

One complication is that what L2 learners and educators perceive as ``grammar'' (such as the one explained in language textbooks) is different from what computer scientists define as grammar (such as the ones used in constituency and dependency parsing)\footnote{We note that the concept of GIs is related to construction grammar~\cite{goldberg2003constructions} and usage-based grammar~\cite{bybee2008usage}, although our approach is agnostic of specific definition of GIs.}. Grammatical items range from simple lexical items such as phrasal verbs to syntactic constructs such as subordinate clauses (Figure \ref{fig:examples}). To find GIs contained in sentences, \citet{ishii2018investigating} wrote regular expressions on sentences automatically tagged with PoS and lemmas.
Grammatical templates~\cite{wang2016grammatical} rely on regular expressions applied to dependency parses. These complex, language-dependent handwritten rules require deep knowledge of computational linguistics to produce, and the algorithms are prone to error propagation from upstream tasks (e.g., PoS tagging and parsing).

Instead, we propose grammatical profiling as {\it span prediction} by borrowing from recent advances in deep NLP methods. Span prediction, widely used for tasks such as reading comprehension~\cite{yu-2018-qanet}, semantic role labeling~\cite{ouchi-etal-2018-span}, and constituency parsing~\cite{stern-etal-2017-minimal,joshi-etal-2018-extending} has several benefits when applied to grammatical profiling:
\begin{itemize}
    \item {\bf Ease of annotation}: annotation does not require linguistic knowledge of any specific tagging or parsing schemes. Annotators only need to mark the start and the end positions of each span and its GI category.
    \item {\bf Partial annotation}: since predictions are made on a per-span basis, the model can learn from partial annotation. Although our datasets are fully annotated, this opens up a wide range of possibilities for leveraging existing resources such as textbooks as (potentially noisy) training signals.
    \item {\bf Overlapping items}: spans can nest and overlap with each other. For example, in the sentence ``I am looking forward to...,'' ``I am looking'' can be tagged as present progressive, and ``looking forward to'' can be tagged as a set phrase. It is difficult, if not impossible, to model overlapping spans with other schemes such as sequential labeling.
\end{itemize}

In this paper, we present GrammarTagger, an open-source grammar profiler based on span prediction (see Figure~\ref{fig:screenshot} for a screenshot). We build a grammar profiler model with $F_1 \approx 0.6$ from only a couple hundred sentences both in English and Chinese. We also show that this performance can be further boosted via multitask and multilingual learning.

As a straightforward application of GrammarTagger, we also present Octanove Learn, a search engine that indexes authentic learning materials by their difficulty and GIs. We do not claim to be the first to build such a search engine—the WERTi system~\cite{meurers-etal-2010-enhancing} analyzes authentic materials such as web pages and highlight grammatical features for learners. They focus on a small set of English GIs (e.g., gerunds, to-infinitives, conditionals) defined using the constraint grammar~\cite{karlsson2011constraint} on top of PoS tagged text. \citet{ott2011information} developed a search engine where users can for texts in terms of their reading difficulties and other linguistic properties. However, little attention has been paid to the grammatical aspects of the materials.

In summary, the contribution of this paper is as follows:
\begin{itemize}
    \item We present a span-based grammatical profiling model and show that a practical grammatical profiler can be built with a small amount of training data, both in English and Chinese.
    \item We show that multitask and multilingual modeling could further improve the performance of grammatical profiling with the same capacity.
    \item We build Octanove Learn, a search engine for language learning materials where users can search for materials by GIs and difficulty.
\end{itemize}

\section{Span-based Grammar Tagging}

\subsection{Model}

\begin{figure}[!t]
\begin{center}
\includegraphics[scale=0.4]{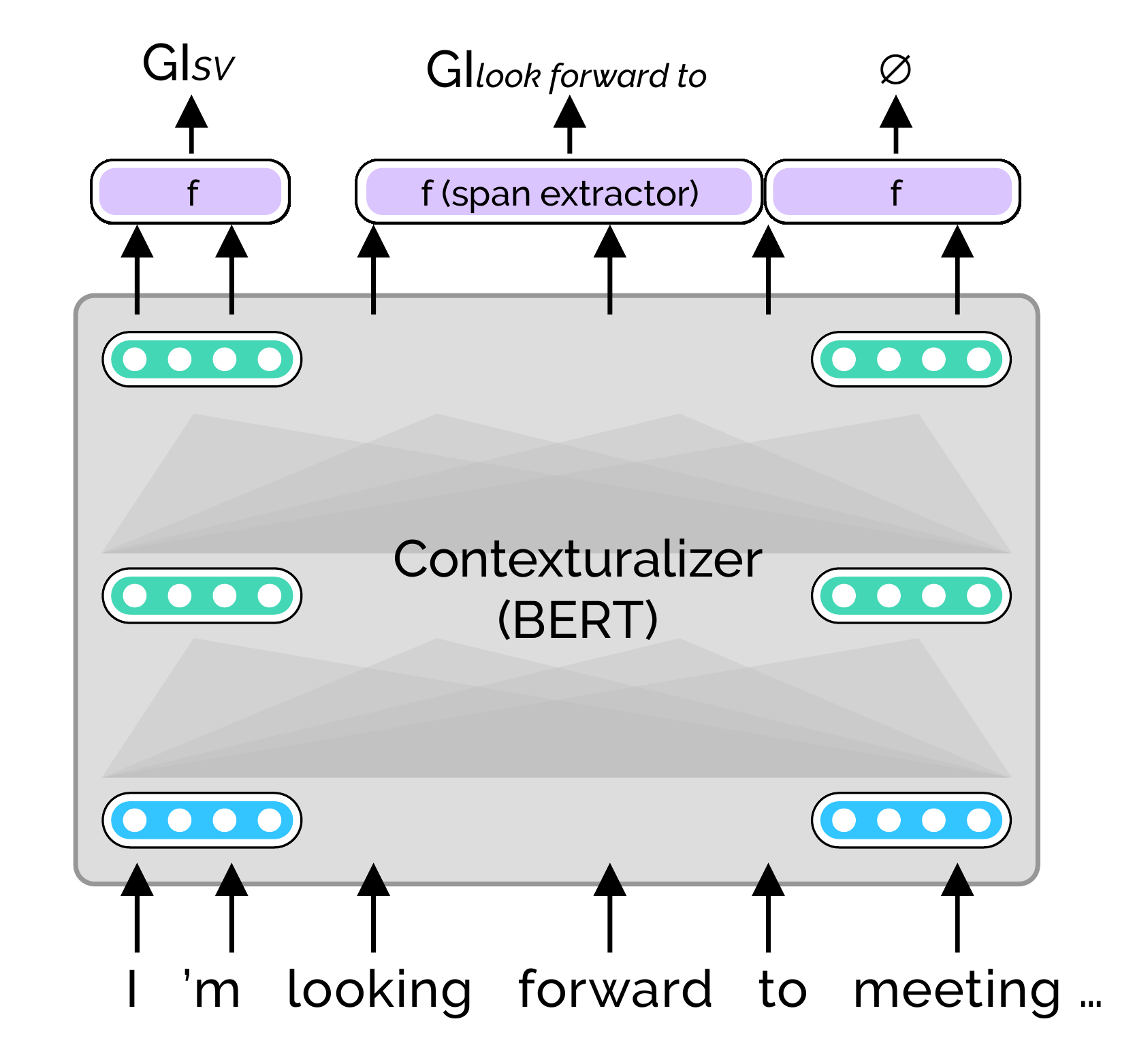} 
\caption{Overview of the proposed architecture. See Section 2.1 for more details.}
\label{fig:model}
\end{center}
\end{figure}

Since GIs are defined as substrings over sentences, span prediction is a natural choice for modeling grammatical profiling, where spans over the input text are classified into distinct GI tags. Our model is heavily inspired by the constituency parsing model proposed in~\citet{joshi-etal-2018-extending}, where they used pretrained language models and span-based partial annotations for domain adaptation of constituency parsers. However, a major difference is that GIs do not need to form a well-formed parse tree and we do not apply a structure prediction step.

Concretely, we use $S({\bf x}) = \{(i, j, t) | 0 \le i \le j < L, i \in \mathbb{Z}, j \in \mathbb{Z}, t \in T\}$ to denote the set of all spans for a given sentence ${\bf x}$ (of length $L$), where $i$, $j$, $t$ are span's start position, end position (inclusive), and its GI tag (which is a member of $T$, the set of all possible tags). Note that conceptually a span is defined for every possible substring of the given sentence, and an empty tag $t = \varnothing$ is used for spans that do not correspond to any GIs. The goal of grammatical profiling is to predict the set of non-empty spans and their GI tags $\tilde S$ for a given sentence.

Our model is a standard span extraction model (Figure~\ref{fig:model}). It first contextualizes the input token sequence ${\bf x} = x_1, ..., x_L$ via BERT~\cite{devlin2019bert}, then extracts span representation via a function $f_{\rm span}$ as follows:

\begin{eqnarray}
    {\bf h}_1, ..., {\bf h}_L &=& \mathtt{BERT}({\bf x}), {\bf h} \in {\mathbb R}^d \nonumber \\
    {\bf h}^s_{i, j} &=& f_{\rm span}((i, j), \{{\bf h}_1, ...\}), {\bf h}^s \in {\mathbb R}^{d_s} \nonumber \\
    \tilde p(t|(i, j)) &=& \mathtt{softmax}(\mathtt{FFN}({\bf h}^s_{i, j})) \nonumber
\end{eqnarray}
where ${\bf h}$ is the contextualized representation obtained by BERT of dimension $d$, ${\bf h}^s$ is the span representation obtained via $f_{\rm span}$, $d_s$ is its dimension, {\tt FFN} is a feedforward network, and $\tilde p(t|(i, j))$ is the predicted probability of the span $(i, j)$ having GI $t$. Finally, we optimize the model using the standard cross entropy loss:
\begin{equation}
    \mathcal{L}_{span} = -\sum_{(i, j, t) \in S({\bf x})} \log \tilde p(t | (i, j))
\end{equation}

We use the concatenation of two endpoints and their difference as the span representation function, i.e., $f_{\rm span}((i, j), \{{\bf h}_1, ..., {\bf h}_L\}) = {\bf h}_i \oplus {\bf h}_j \oplus ({\bf h}_i - {\bf h}_j)$ where $\oplus$ is vector concatenation.

\subsection{Datasets}

\begin{table}[t]
\centering
\begin{tabular}{lrrr}
\toprule
Language                     & \# Passages  & \# Sents   & \# GIs \\ \midrule
English                      & 30    & 415   & 3,466  \\
Chinese                      & ---   & 558   & 1,730  \\ \bottomrule
\end{tabular}
\caption{Dataset statistics}
\label{tab:dataset}
\end{table}

We used an in-house, difficulty-controlled English dataset—it consists of 5 passages  per CEFR (the Common European Framework of Reference)~\cite{council2001} level written by a native speaker with ESL background. The Chinese dataset consists of level-balanced reading passages taken from HSK (a standard Chinese proficiency test) sample questions\footnote{\url{http://www.chinesetest.cn/godownload.do}}. Both datasets were annotated with GIs by native speakers of each language with linguistics background. For English, we used CEFR-J Grammar Profile~\cite{ishii2018investigating}, a list of GIs and their per-level statistics developed by the CEFR-J project~\cite{negishi2013progress}. For Chinese, we used a list of GIs annotated with HSK levels generously provided by Zero to Hero Education\footnote{\url{https://www.zerotohero.ca/}}. Both lists are published at Open Language Profiles\footnote{\url{https://www.openlanguageprofiles.org/}} under a permissive creative commons license. Table~\ref{tab:dataset} shows the statistics of the datasets we used for training and evaluating the model.

Note that some GIs correspond to {\it lack} of some linguistic units—this happens for, e.g., omission of the subordinate clause marker ``that'' in English and omission of the possessive marker ``de'' in Chinese. For these GIs, we annotated the subsuming structures, e.g., the main clause (see the third example in Figure~\ref{fig:examples}) or the entire noun phrase, with the corresponding GI.

\subsection{Evaluation Metrics}

We use the following metrics for model evaluation:

\begin{itemize}
    \item Labeled precision ($P_L$) and recall ($R_L$) defined as follows, and their $F_1$ measure:
$$
    P_L = \frac{|S_L \cap \tilde S_L|}{|\tilde S_L|}, \quad R_L = \frac{|S_L \cap \tilde S_L|}{|S_L|},
$$
    where $S_L$ and $\tilde S_L$ are the sets of {\it labeled} (i.e., $(i, j, t)$) non-empty spans in the ground truth and the model prediction, respectively.
    \item Unlabeled precision and recall defined as follows, and their $F_1$ measure:
$$
    P_U = \frac{|S_U \cap \tilde S_U|}{|\tilde S_U|}, \quad R_U = \frac{|S_U \cap \tilde S_U|}{|S_U|},
$$
    where $S_U$ and $\tilde S_U$ are the sets of {\it unlabeled} (i.e., $(i, j)$) non-empty spans in the ground truth and the model prediction, respectively.

    \item Macro-averaged precision ($P_M$) and recall ($R_M$) defined as follows, and their $F_1$ measure:
\begin{eqnarray}
    P_t = \frac{|S_t \cap \tilde S_t|}{|\tilde S_t|},& &R_t = \frac{|S_t \cap \tilde S_t|}{|S_t|} \nonumber \\
    P_M = \frac{1}{|T|} \sum_{t \in T} P_t,& &R_M = \frac{1}{|T|} \sum_{t \in T} R_t, \nonumber
\end{eqnarray}
    where $T$ is the set of all distinct GI tags, $S_t$ and $\tilde S_t$ are the sets of spans for tag $t$ in the ground truth and the model prediction, respectively.
\end{itemize}

We used the labeled $F_1$ measure as the main evaluation metric for tuning the model parameters.

\subsection{Experiments}

The feedforward network is a single linear layer, initialized with Xavier initialization~\cite{glorot-2010-understanding}. We used {\tt bert-base-cased} for English and {\tt bert-base-chinese} for Chinese from the Transformers library~\cite{wolf-etal-2020-transformers} as contextualizers. The sentence was tokenized via the tokenizers corresponding to the pretrained model used. This means that Chinese text was effectively tokenized into individual characters. We implemented the model using AllenNLP~\cite{gardner-etal-2018-allennlp}.

Only spans of up to 30 tokens long were considered for prediction, and GI tags that appear only once in the dataset were collapsed to a special UNK tag. We truncated long sentences containing more than 128 tokens.

We used a batch size of 50, and the Adam optimizer~\cite{kingma2015adam} with a learning rate of $1\times10^{-4}$, $\beta_1 = 0.9$, $\beta_2 = 0.999$, and $\epsilon = 1\times10^{-8}$. We trained the model for 100 epochs and chose the best model based the validation labeled $F_1$ measure. We conducted 10-fold cross validation where we chose one fold for testing, another for validation, and the rest for training, reporting the averaged metrics on the test portion.

\begin{table}[t]
\centering
\begin{tabular}{llrrr} \toprule
Lang &  & \multicolumn{1}{l}{Prec} & \multicolumn{1}{l}{Rec} & \multicolumn{1}{l}{$F_1$} \\ \midrule
\multirow{3}{*}{en} & Labeled & 0.628 & 0.489 & 0.549 \\
 & Unlabeled & 0.694 & 0.540 & 0.606 \\
 & Macro & 0.205 & 0.188 & 0.187 \\ \midrule
\multirow{3}{*}{zh} & Labeled & 0.731 & 0.456 & 0.560 \\
 & Unlabeled & 0.748 & 0.466 & 0.573 \\
 & Macro & 0.150 & 0.142 & 0.141 \\ \bottomrule
\end{tabular}
\caption{Grammatical profiling performance for English (en) and Chinese (zh)}
\label{tab:result}
\end{table}

Table~\ref{tab:result} shows the main result. The model achieves a decent level of grammatical profiling performance in spite of the small training dataset, suggesting that our method is effective for bootstrapping a practical grammar profiler. Note that the macro-averaged precision, recall, and $F_1$ measure are low, which is due to the heavily skewed distribution of GI tags. A small number of GI tags such as pronouns and tense markers occur frequently while others do not, which makes it difficult to correctly predict infrequent GIs from small training data. Robustly predicting such rarely occurring GIs is future work.

\section{Multitask and Multilingual Learning}

\subsection{Combining with Readability Assessment}

Since GIs can be useful features for readability/difficulty assessment, it is natural to ask if grammatical profiling and readability assessment can be solved jointly and if these two tasks benefit from each other. As a proof of concept, we built a simple multitask model where the {\tt [CLS]} embeddings from the final layer of BERT were fed to a linear layer then a softmax layer to predict the difficulty (one of six CEFR levels) of the input. The classification head is trained with standard cross entropy, which we denote $\mathcal{L}_{level}$. The final loss used for the joint model is:
\begin{eqnarray}
    \mathcal{L} = \mathcal{L}_{span} + \alpha\mathcal{L}_{level}
\end{eqnarray}

We ran minimal grid search to find the optimal value $\alpha$ on the validation set. The result is shown in Table~\ref{tab:multitask}.

\begin{table}[t]
\centering
\begin{tabular}{lrrrr} \toprule
Model & \multicolumn{1}{l}{\small{Labeled}} & \multicolumn{1}{l}{\small{Unlabeled}} & \multicolumn{1}{l}{\small{Macro}} & \multicolumn{1}{l}{\small{Acc.}} \\ \midrule
en-single & 0.549 & 0.606 & 0.187 & --- \\
en-multi & 0.552 & 0.605 & 0.179 & 0.561 \\ \midrule
zh-single & 0.560 & 0.573 & 0.141 & --- \\
zh-multi & 0.556 & 0.566 & 0.140 & 0.557 \\ \bottomrule
\end{tabular}
\caption{Grammatical profiling performance (labeled $F_1$, unlabeled $F_1$, and macro $F_1$) combined with readability assessment accuracy}
\label{tab:multitask}
\end{table}

Compared to single models, multitasking with readability assessment improved the performance of grammatical profiling for English, while slightly hurting it for Chinese, although the effect is limited for both languages. The additional task may have helped regularize the contextualizer and offered a ``priming'' effect for grammatical profiling. Note that readability is usually estimated with longer texts such as paragraphs, which may have contributed to the suboptimal results shown here. We leave more detailed analysis and search for better joint/pipeline architectures for future work. 

\subsection{Multilingual Grammar Profiling}

\begin{table}[t]
\centering
\begin{tabular}{llrrr} \toprule
Lang & Model & \multicolumn{1}{l}{\small{Labeled}} & \multicolumn{1}{l}{\small{Unlabeled}} & \multicolumn{1}{l}{\small{Macro}} \\ \midrule
\multirow{2}{*}{en} & en & 0.549 & 0.606 & 0.187 \\
                 & en+zh & 0.565 & 0.619 & 0.211 \\ \midrule
\multirow{2}{*}{zh} & zh & 0.560 & 0.573 & 0.141 \\
                 & en+zh & 0.566 & 0.588 & 0.147 \\ \bottomrule
\end{tabular}
\caption{Grammatical profiling performance (labeled $F_1$, unlabeled $F_1$, and macro $F_1$) in a mono- and multi-lingual settings}
\label{tab:multilingual}
\end{table}

Little attention has been paid to building multilingual models for educational applications, except for e.g., \cite{vajjala2018experiments}. By definition, grammatical profiling is a language dependent task, and past studies dealt only with individual (usually high-resource) languages. However, being based on a language independent architecture, our model opens up a whole new set of possibilities for training multilingual grammar profilers. To investigate its generalization ability in a multilingual setting, we trained an English+Chinese joint model by combining the two training datasets and by using the {\tt bert-base-multilingual-cased} model as the contextualizer. The set of target GI tags is extended to the union of all the tags in both languages. To the best of our knowledge, ours is the first model that solves grammatical profiling in multiple languages.

Table~\ref{tab:multilingual} shows the comparison between mono- and multi-lingual settings. The performance of the multilingual model is not just changed—it in fact improves grammatical profiling for English and Chinese, even though the model needs to encode the information from both languages with almost the same capacity. This suggests multilingual modeling is a promising venue for this particular task, being potentially beneficial for low-resource, closely related languages, although detailed investigation of multilingual models is future work. 

Interestingly, the improvement in {\it unlabeled} $F_1$ measures between the mono- and multi-lingual models is larger than that in labeled measures for Chinese. We posit that, even between vastly different languages such as English and Chinese, {\it where} grammatically salient constructs occur in sentences might have a lot more in common across languages than {\it what} these constructs look like. 

\begin{figure}[!t]
\begin{center}
\includegraphics[scale=0.21]{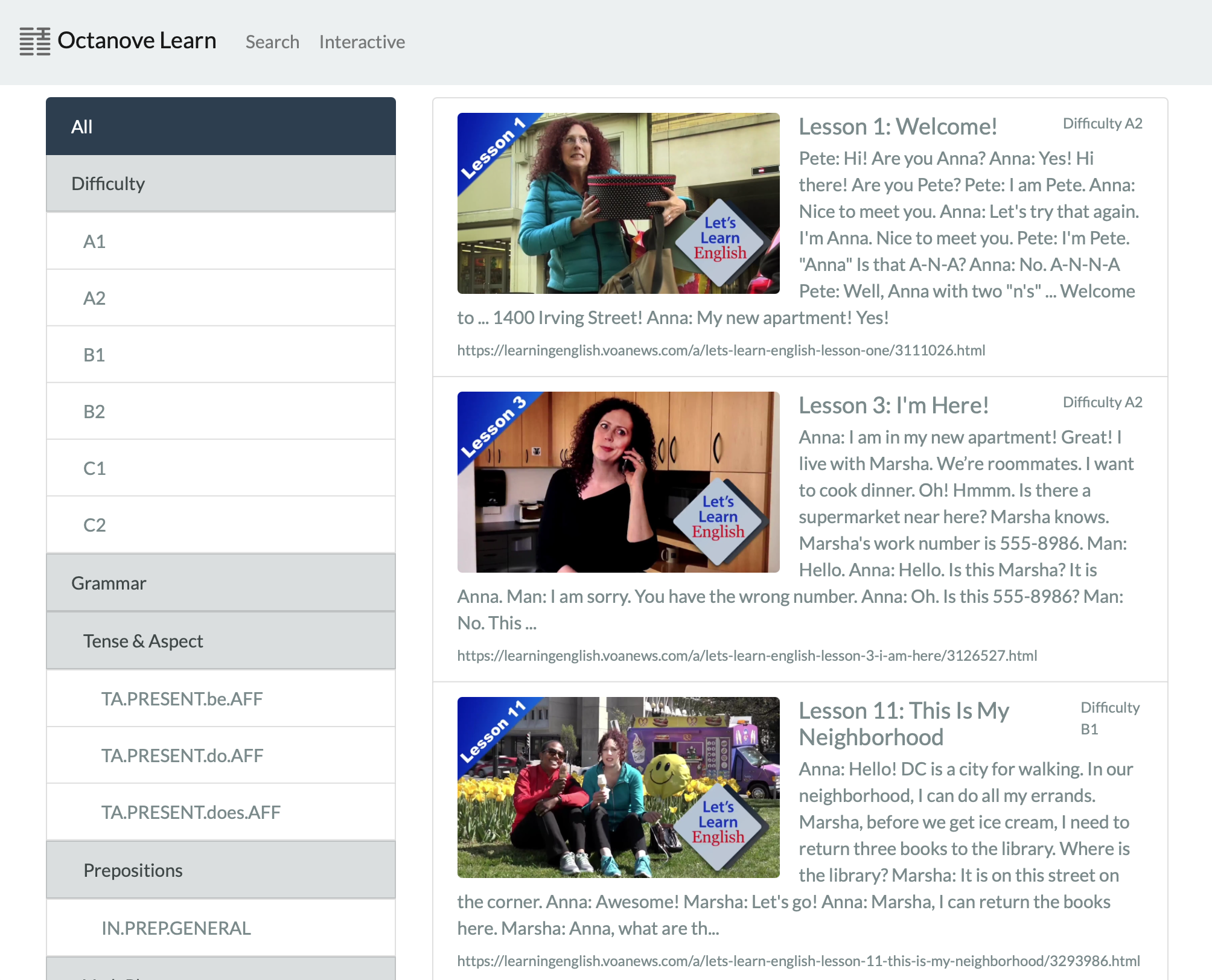} 
\caption{Screenshot of Octanove Learn. Language learners and educators can search examples of GIs with specific difficulties.}
\label{fig:learn}
\end{center}
\end{figure}

\section{Searching Materials by Grammar}

For our demo application we present Octanove Learn\footnote{\url{http://learn.octanove.com/}} (Figure~\ref{fig:learn}), a search engine which indexes learning materials by the GIs they include and their difficulty, as predicted by GrammarTagger. Here, we define individual materials to each be a document $X$ consisting of a set of sentences $x$, where the tags associated with a document are the union of the span labels provided by GrammarTagger for all its sentences, and its difficulty is the most common difficulty provided by GrammarTagger among all its sentences. Let $G(x)$ and $D(x)$ be the set of GIs and the difficulty of sentence $x$ returned by GrammarTagger. The set of GIs and the difficulty of document $X$ are defined as:
\begin{eqnarray}
    G(X) &=& \bigcup\limits_{x \in X} G(x) \nonumber \\
    D(X) &=&  \argmax\limits_{d} |{\{x \in X \,| D(x) = d\}}| \nonumber
\end{eqnarray}

We can then retrieve the set of documents including a given GI $g$ as $\{X \in {\mathcal X} \,|\, g \in G(X)\}$, and documents for a given difficulty $d$ as $\{X \in {\mathcal X} \,|\, D(X) = d \}$, where ${\mathcal X}$ is the collection of all documents. It would also be trivial to implement the the ability to search for specific sentences by their GIs and difficulty in a similar fashion.

We imagine the primary application of this technology to be indexing existing data so that it can be searched by language learners and/or educators looking for examples of GIs at specific difficulties. However, we also include interactive profiling functionality which can be used to identify GIs in arbitrary input sentences, both to demonstrate the functionality of GrammarTagger and because it could be helpful to learners attempting to identify GIs in sentences they are having difficulty understanding.

\section{Conclusion}

We presented GrammarTagger, a grammar profiler that identifies grammatical items from text based on simple and flexible span prediction. The experiments showed that the model achieved the grammatical profiling performance of $F_1 \approx 0.6$ from only a couple hundred sentences both in English and Chinese. We also show that this performance can be further boosted via multitask and multilingual learning.

We are planning on extending GrammarTagger to other languages than English and Chinese. Also, as we've shown partially, multitask and multilingual learning are a promising venue for building a more robust and better grammatical profiling model and we leave the investigation as future work. 

\section*{Acknowledgments}

The authors would like to thank Prof. Yasutake Ishii at Seijo University and Jon Long at Zero to Hero Education for their permission to use the grammatical resources. We also thank Kathleen Hall, Xiaoyue Peng, and Thomas Stones for their help with annotation.


\bibliographystyle{acl_natbib}
\bibliography{anthology,acl2021}


\end{document}